\documentclass{elsart}
\usepackage{graphicx}
\begin{document}
\begin{frontmatter}
\title{Coding Facial Expressions\\with Gabor Wavelets\\(IVC Special Issue)}
\date{September, 1998}
\author[ATR]{Michael J. Lyons\thanksref{CA}},
\author[ATR]{Miyuki Kamachi},
\author[Tohoku]{Jiro Gyoba}
\address[ATR]{ATR Human Information Processing Research Laboratory,2-2 Hikaridai, Seika-cho,
Soraku-gun, Kyoto 619-0288, Japan}
\address[Tohoku]{Psychology Department, Tohoku University, Sendai, 980-8576,
Japan}
\thanks[CA]{This manuscript is a modified version of a conference article \cite{FG98}, that was invited for publication in a special issue of Image and Vision Computing dedicated to a selection of articles from the IEEE Face \& Gesture 1998 conference. The special issue never materialized. MJL is now with Ritsumeikan University.}
\begin{abstract}
We present a method for extracting information about facial expressions
from digital images.The method codes facial expression images 
using a multi-orientation, multi-resolution set of Gabor
filters that are topographically ordered and approximately aligned
with the face. A similarity space derived from this code
is compared with one derived from semantic ratings of the images
by human observers. Interestingly the low-dimensional structure of the image-derived
similarity space shares organizational features with the circumplex model of affect, suggesting a bridge between categorical and dimensional representations of facial expression.
Our results also indicate that it would be possible to construct
a facial expression classifier based on a topographically-linked 
multi-orientation, multi-resolution Gabor coding of the facial images
at the input stage. The significant degree of psychological plausibility exhibited by the proposed
code may also be useful in the design of human-computer interfaces.
\end{abstract}
\end{frontmatter}
\section{Introduction}
Processing of information related to social relationships in groups is
an important computational task for primates. The recognition of
kinship, identity, sex and emotional or attentive state of an
individual from the appearance of the face are all examples of
this type of visual task (for a review see \cite{VB}). Whether or not
we are explicitly conscious of it, such non-verbal information channels
are a critical component of human communication. It would be desirable
to make use of these modes for human-compute interaction or computer-mediated human-human interaction.  The development of computational methods for handling face 
and gesture information is a critical step to achieve this goal.

The current paper concentrates on the representation of facial
expressions. The face displays several classes of perceptual cues to emotional state: relative displacements of features (opening the mouth), quasi-textural changes in the skin surface (furrowing the brow), and changes in skin hue (blushing); and the time course of these
signals.  The methods presented in this paper treat feature displacements and quasi-textural cues.  Motion is considered only implicitly through the comparison of images. We do not examine colour information.

Our general framework for representing facial expressions uses
topographically ordered, spatially localized filters to code
patterns in the images. The filters consist of a multi-resolution,
multi-orientation bank of Gabor wavelet functions. A similar
representation appears in the automatic face recognition system
developed by the von der Malsburg group \cite{Lades}.

Previous work on automatic facial expression processing includes
studies using representations based on optical flow estimation from
image sequences \cite{Mase,Yacoob,Bart}; principal components
analysis of single images \cite{Cottrell,Bart}; and physically-based
models \cite{Essa}.  This paper describes the first study to use
Gabor wavelets to code facial expressions.  Our findings indicate that
it is possible to build an automatic facial expression recognition system based on a Gabor wavelet code that has a significant level of
psychological plausibility. The recently obtained results
of Zhang et al. \cite{Zhang} support this by demonstrating expression
classification using Gabor coding and a multi-layer perceptron.

This work is the first to use Gabor wavelets to code facial
expressions.\footnote{Preliminary reports on the research were presented at the ARVO'97 conference in May 1997 \cite{ARVO97}, and at a workshop in Okinawa in June 1997 \cite{OK97}.}  Our approach also differs from previous studies on
expression recognition in that we test the ``fidelity'' of the facial
expression representation scheme: if two facial expressions are
perceived as being similar by human observers, they should be
neighbours in the space of the representation. In addition to being a
potentially important engineering criterion in the design of facial
expression processing systems, fidelity is convenient in testing the utility of a representation because it allows the use of examples of
expression images that are not pure (or even standard) facial
expressions. Rather than assigning training examples to hard
expression categories and testing the classification performance of a
model, we can examine the extent to which the representation model
reflects human judgements on the expression content of the face.

\begin{figure}[t]
\begin{center}
\includegraphics[width=.8\linewidth]{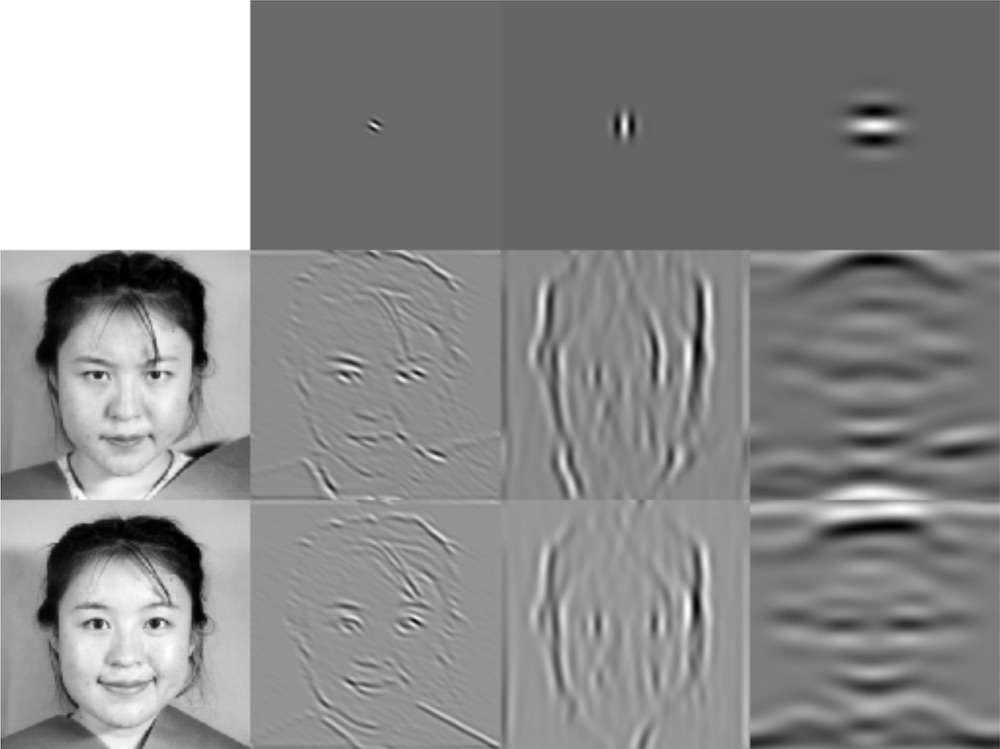}
\caption{Examples of Gabor filter responses to two facial expression images 
for three of the filters used.}
\label{Gabor}
\end{center}
\end{figure}
\section{Multi-Scale, Multi-Orientation Gabor Coding}
To extract information about facial expression, each 256 by 256-pixel image,
$I$, was convolved with a multiple spatial resolution,
multiple orientation set of Gabor filters (Fig.1),
$G_{\vec{k},+}$ and $G_{\vec{k},-}$. The sign subscript indicates
filters of even and odd phase, while $\vec{k}$, the filter wave-vector,
determines the spatial frequency and orientation tuning of the
filter. A description of the complex-valued two dimensional Gabor
transform is given by Daugman \cite{Daug85}. Responses of the filters
to the image were combined into a vector, $\bf{R}$, with components
given by:
\begin{equation}
R_{\vec{k},\pm} ( \vec{r}_{0} ) = \int G_{\vec{k},\pm} (\vec{r}_{0}, \vec{r})
I(\vec{r}) d \vec{r},
\end{equation}
where,
\begin{equation}
G_{\vec{k},+}(\vec{r}) = \frac{k^2}{\sigma^2} e^{-k^2 \|\vec{r} - \vec{r}_{0}\|^2/2 \sigma ^2}
cos(\vec{k} \cdot (\vec{r} - \vec{r}_{0})) - e^{-\sigma ^2 /2}),
\end{equation}
\begin{equation}
G_{\vec{k},-}(\vec{r})  = \frac{k^2}{\sigma^2} e^{-k^2 \|\vec{r} - \vec{r}_{0}\|^2/2 \sigma ^2}
sin(\vec{k} \cdot (\vec{r} - \vec{r}_{0})).
\end{equation}
The integral of the cosine Gabor filter, $e^{- \sigma ^2 / 2}$, is
subtracted from the filter to render it insensitive to the absolute
level of illumination.  The sine filter does not depend on the
absolute illumination level.  Three spatial frequencies were used with
wave-numbers:
\begin{equation}
k = \{ \frac{\pi}{2},\frac{\pi}{4},\frac{\pi}{8} \}
\end{equation}
measured in inverse pixels. The highest frequency is set at half the
Nyquist sampling frequency, with frequency levels spaced at octaves;
$\sigma = \pi $ was used in all calculations, giving a filter
bandwidth of about an octave, independent of the frequency level. Six
wave-vector orientations were used, with angles equally spaced at
intervals of $\frac{\pi}{6}$ from $0$ to $\pi$.

The components of the Gabor vector, $R_{\vec{k}}$, are defined as the
amplitude of the combined even and odd filter responses:
 \begin{equation}
 R_{\vec{k}} =
\sqrt{R^{2}_{\vec{k},+}+R^{2}_{\vec{k},-}}
\end{equation}
The response amplitude is
less sensitive to position changes than are the linear filter responses.
\begin{figure}[t]
\begin{center}
\includegraphics[width=.8\linewidth]{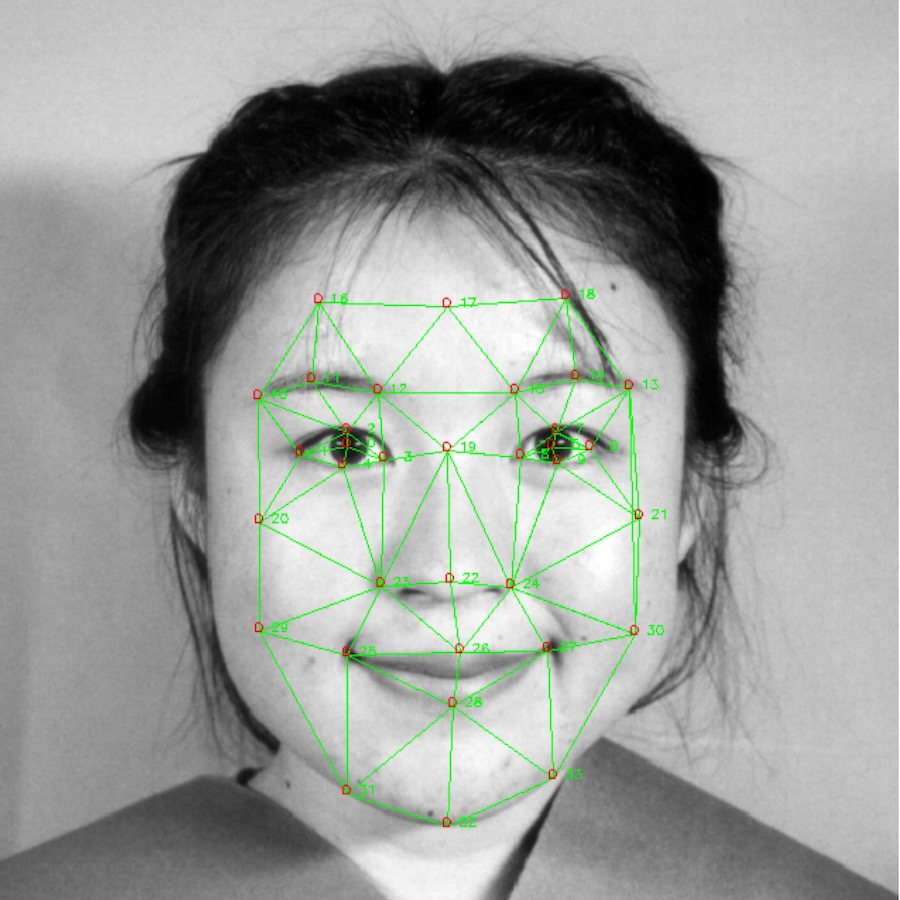}
\caption{The 34 node grid used to represent facial geometry.}
\label{Grid}
\end{center}
\end{figure}
To study the similarity space of Gabor coded facial images, we compared responses of filters having the same spatial frequency and orientation
preference at corresponding points in the two facial
images. We use the normalized dot product to quantify the similarity
of two Gabor response vectors. We calculate the similarity of two facial images as the average of the Gabor vector similarity over all
corresponding facial points. Since Gabor vectors at neighbouring pixels
are strongly correlated, it is sufficient to carry out this calculation at points on a sparse grid covering the face (Fig. 2). The automatic face recognition system developed by the von der Malsburg group \cite{Lades} uses a related similarity measure. However, the filter parameters used here differ from those used in that work.  Previous work has demonstrated automatic systems for scaling the face and registering a graph approximately with the features of the face \cite{Lades}.  For this reference study, the highest precision positioning was desirable. Therefore grids were positioned manually on facial images scaled to a standard size.
\section{Facial Expression Dataset}
\begin{figure}[t]
\begin{center}
\includegraphics[width=.7\linewidth]{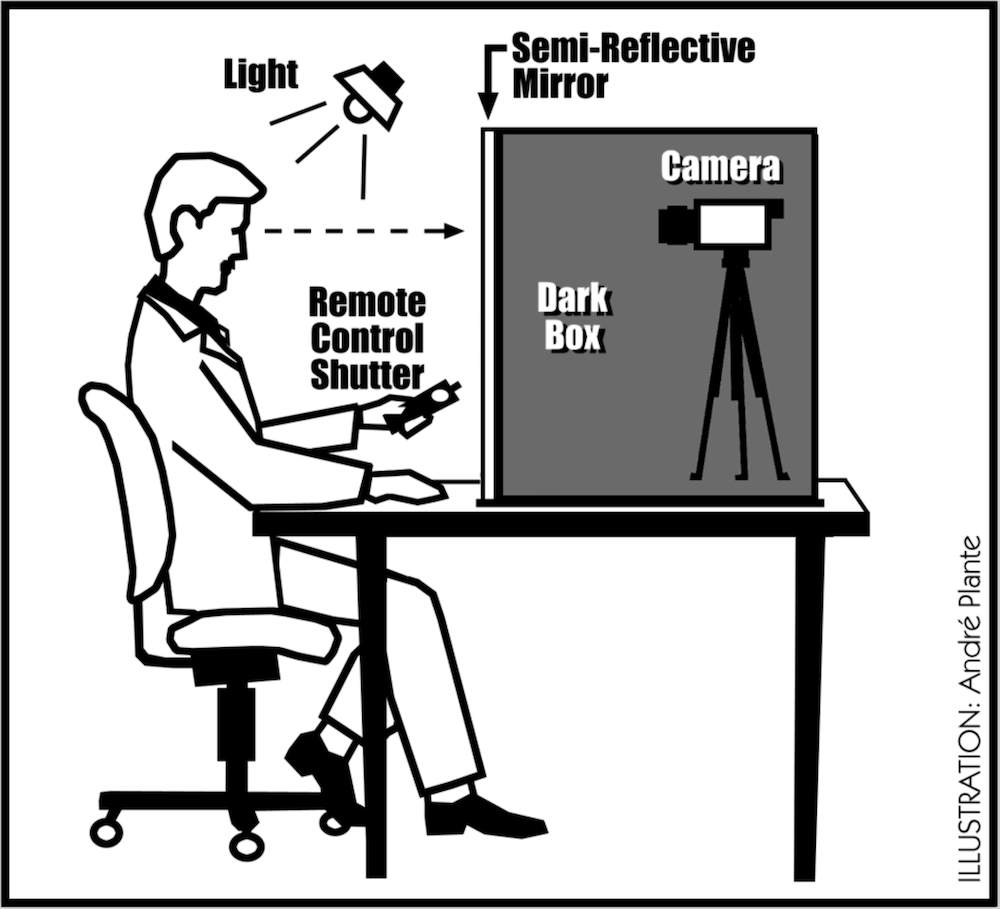}
\caption{Apparatus used to photograph facial expressions.}
\label{setup}
\end{center}
\end{figure}
A dataset of facial expression images was collected.  Ten expressers
posed 3 or 4 examples of each of the six basic facial expressions (happiness, sadness, surprise, anger, disgust, fear) \cite{Ekman} and a neutral face for a total of 219 images of facial expressions. To simplify the
experimental design, only Japanese female expressers and subjects were
employed. Figure 3 shows the apparatus used to photograph the
expressers. Each expresser took pictures of
herself while looking through a semi-reflective plastic sheet towards
the camera. Hair was tied away from the face to expose all
expressive zones of the face.  We positioned tungsten lights to
illuminate the face evenly. A box enclosed the region between
the camera and plastic sheet to reduce back-reflection. The images were printed as monochrome photographs and digitized using a flatbed scanner. Figure 4 shows sample images. 
\begin{figure}[t]
\begin{center}
\includegraphics[width=\linewidth]{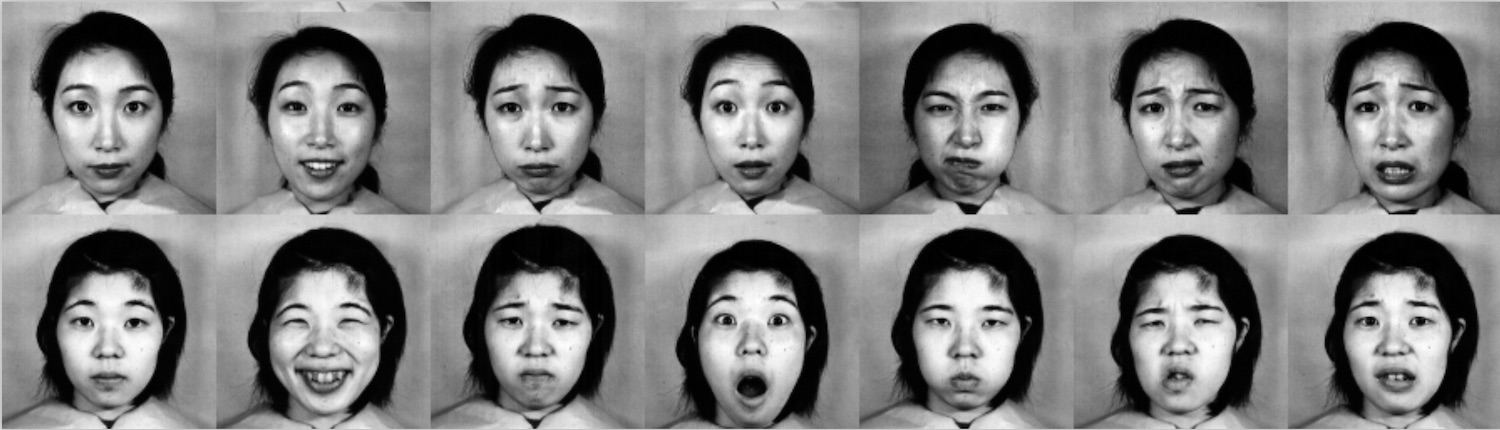}
\caption{Examples of images from the facial expression dataset.}
\label{Faces}
\end{center}
\end{figure}
\section{Semantic rating of facial expression images.}
To provide a basis for testing the fidelity of the Gabor
representation, we directly compare the similarities as measured from the  Gabor coded images and derived from human judgements. With this procedure, we do not have to use the expression labels attached to each image (the emotion posed by that the expresser) when comparing the model with the data. Instead, viewers rate the emotional content of each image using emotion adjectives.  This approach captures variations in intensity and blends of mixed facial expressions and reduces the epistemological difficulties of working with photographs of the Ekman standard basic facial expressions in a different cultural context.

Experimental subjects rated pictures for the degree of each component
basic expression on a five-point Likert scale.  A total of 92 Japanese female undergraduates took part in the study.  The subject pool was divided into four groups: 1.A, 1.B, 2.A, and 2.B.  Group 1.A (31 subjects)
rated 108 pictures on six basic facial expressions (happiness,
sadness, surprise, anger, disgust and fear). Group 1.B (31 subjects)
rated the complementary set of 111 pictures (out of the 219 total) on
the six basic expressions. Both Group 1.A and 1.B saw images of all
seven expression categories (including fear images).  Group 2.A (15
subjects) rated 94 pictures on five of the six basic facial
expressions (fear was excluded). Group 2.B (15 subjects) rated a
different set of 93 images on the five basic facial expressions (fear
excluded). The images presented to Group 2.A and 2.B excluded fear
expressions. Each image was thus labelled with a 5 or 6 component
vector with ratings averaged over all subjects. Similarities between
these semantic vectors were calculated using the Euclidean distance.

In pilot experiments, we found that fear ratings showed greater variability than ratings for the other expression categories. For this reason, we also ran a set of experiments that excluded pictures of fear expressions and fear ratings.

\section{Results}
Facial expression image similarity computed using the Gabor coding and
semantic similarity computed from human ratings were compared by rank
correlation. It is convenient to compare similarity spaces rather than
categorization performance as this avoids the problem that posed
expressions are not necessarily pure examples of a single expression category.

As a control, geometric similarity was also rank correlated with the
semantic ratings similarity values. The distance of each grid point
(Fig. 2) from the point at the nose tip formed the components of a 33
dimensional shape vector. Dissimilarity between two grid
configurations were calculated using the Euclidean distance.
\begin{table}
\begin{center}
\begin{tabular}{|c|c|c|} \hline\hline
Expresser Initials & Gabor & Geometry \\ \hline
KA & 0.593 & 0.467 \\ 
KL & 0.465 & 0.472 \\ 
KM & 0.616 & 0.527 \\ 
KR & 0.636 & 0.368 \\ 
MK & 0.472 & 0.287 \\ 
NA & 0.725 & 0.358 \\ 
NM & 0.368 & 0.099 \\ 
TM & 0.423 & 0.282 \\ 
UY & 0.648 & 0.074 \\ 
YM & 0.538 & 0.455 \\ \hline
Average & 0.568 & 0.366 \\ 
\hline\hline 
\end{tabular}
\end{center}
\caption{Rank correlation between model and semantic rating similarities.}
\label{Ftab}
\end{table}
For the experiments in which all facial expressions were included
(i.e.  comparison with data from subject groups 1.A and 1.B) the rank
correlation between Gabol model and human data ranged from 0.42
(expresser TM) to 0.725 (expresser NA) with an average value of 0.568.
For the geometry based control, rank correlation between model and
data ranged from 0.074 (expresser UY) to 0.527 (expresser KM) with an
average value of 0.366. Correlation results for all expressers are
listed in Table 1.  With fear stimuli and ratings excluded (data from
subject groups 2.A and 2.B) the rank correlation between Gabor model
and data ranged from 0.624 (expresser TM) to 0.782 (expresser KA),
with an average value of 0.679. For the geometry based control, rank
correlations between model and data ranged from 0.206 (expresser UY)
to 0.619 (expresser KM) with an average value of 0.462. Correlation
results for all expressers are listed in Table 2.
\begin{table}
\begin{center}
\begin{tabular}{|c|c|c|} \hline\hline
Expresser Initials & Gabor & Geometry \\ \hline
KA & 0.782 & 0.574 \\ 
KL & 0.634 & 0.500 \\ 
KM & 0.744 & 0.619 \\ 
KR & 0.684 & 0.401 \\ 
MK & 0.644 & 0.512 \\ 
NA & 0.696 & 0.420 \\ 
NM & 0.458 & 0.207 \\ 
TM & 0.624 & 0.425 \\ 
UY & 0.653 & 0.206 \\ 
YM & 0.650 & 0.506 \\ \hline
Average & 0.679 & 0.462 \\ 
\hline\hline 
\end{tabular}
\end{center}
\caption{Rank correlations between model and semantic rating similarities
for experiments which excluded fear stimuli and ratings.}
\label{Ntab}
\end{table}
Expresser NM was considered to be an outlier and excluded from the
above quoted averages and ranges. On closer inspection NM's expressions appeared to be difficult to interpret.

All rank correlations quoted were calculated using Spearman's
method. The two sided significance of all of the deviation of all rank
correlations calculated indicated a high level of significance.  In
all cases the correlation coefficient was greater for the Gabor model
than for the model based solely on geometric displacement of feature
points.
\begin{figure}[t]
\begin{center}
\includegraphics[width=.49\linewidth]{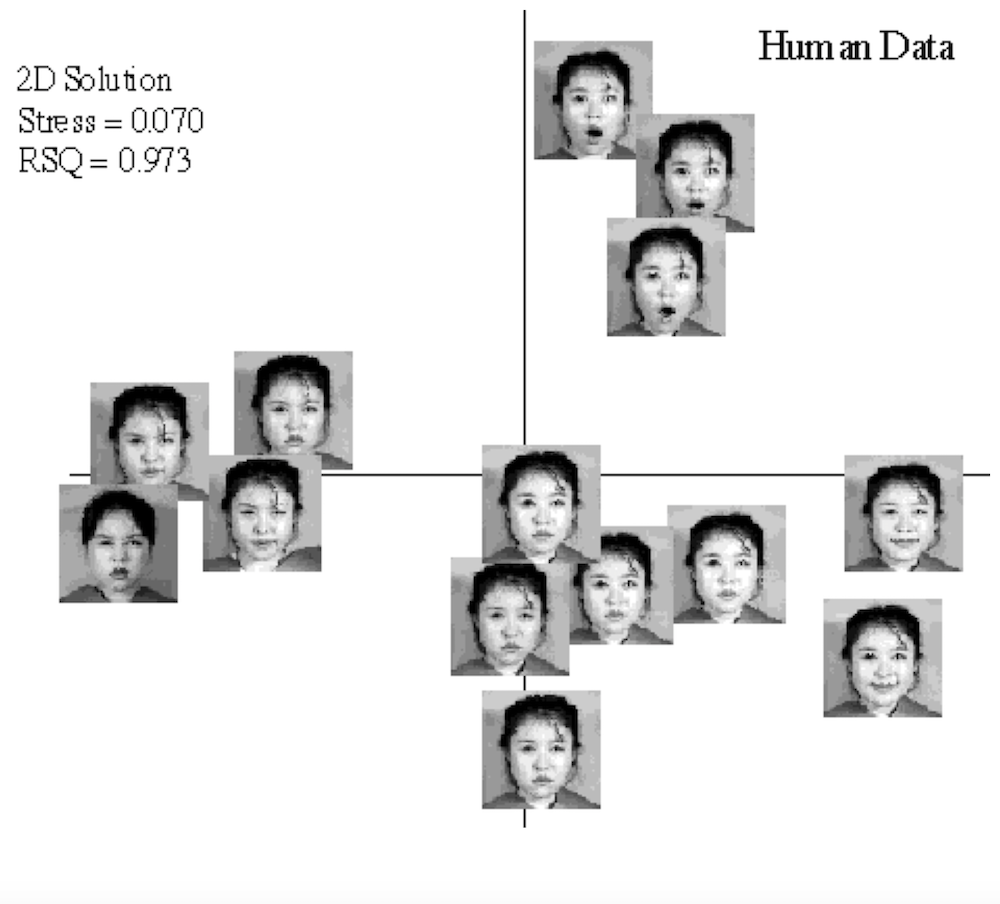}
\includegraphics[width=.49\linewidth]{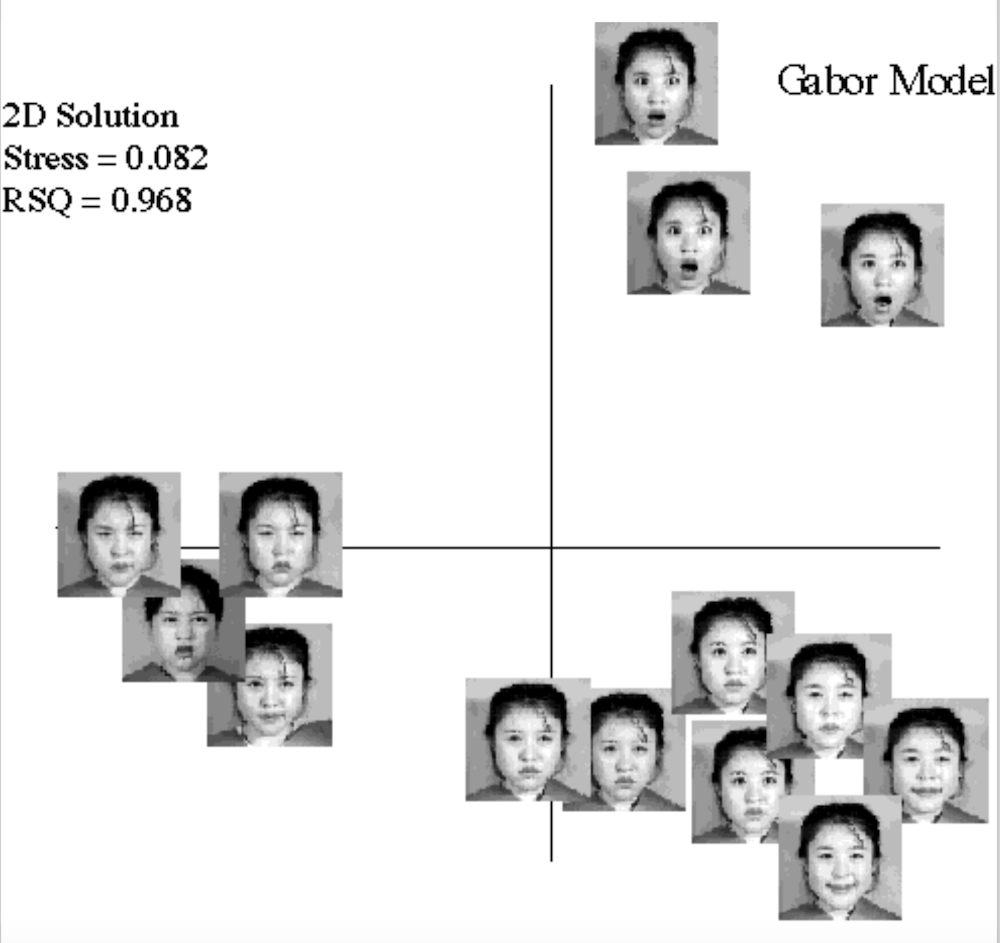}
\caption{nMDS solutions for Gabor and semantic rating similarities.}
\label{Fig5}
\end{center}
\end{figure}
Gabor and human similarity data was analyzed using non-metric
multidimensional scaling (nMDS) using the ALSCAL algorithm
\cite{ALSCAL}.  nMDS embeds points in a Euclidean space in such a way
that the distances between points preserves the rank order of the
dissimilarity values between those points. ``Stress'' and ``Rsq''
respectively measure the residual misfit of the Euclidean distance to
the dissimilarities and the squared correlation between distances and
dissimilarities.  By monitoring these parameters as the number of nMDS
dimensions was increased, it was found that two dimensions provide an
adequate embedding of the similarity data. Figs. 5, 6, and 7 show
sample nMDS solutions for human ratings similarity values and Gabor
code derived similarity values.  In figs. 6 and 7, the following
abbreviations are used: NE - Neutral, HA - Happiness, SA - Sadness,
SU-Surprise, AN - Anger, DI - Disgust.  Fig. 5 shows sample
nMDS solution in which images have been positioned at their
coordinates in the Euclidean space.
\begin{figure}[b]
\begin{center}
\includegraphics[width=.49\linewidth]{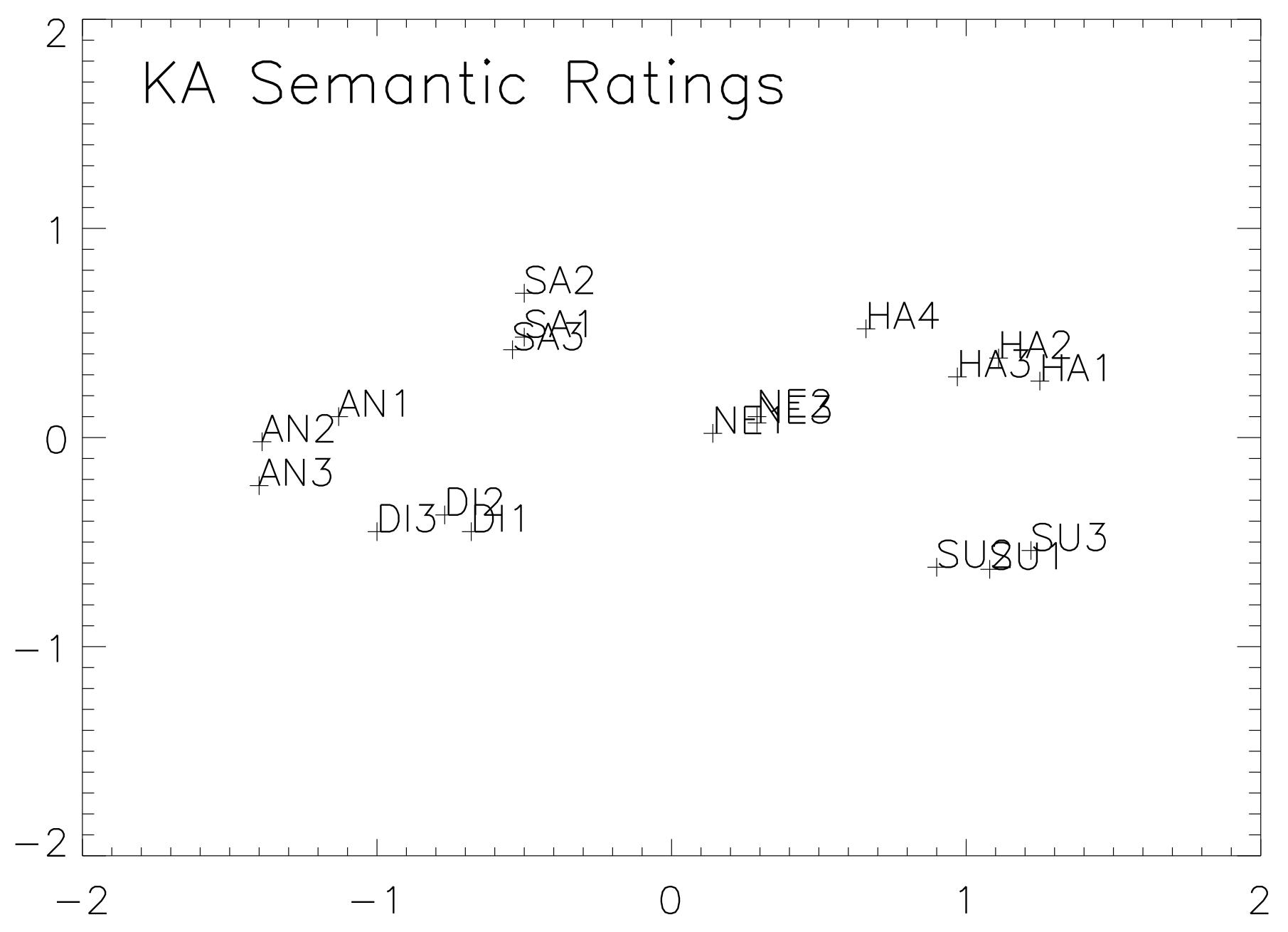}
\includegraphics[width=.49\linewidth]{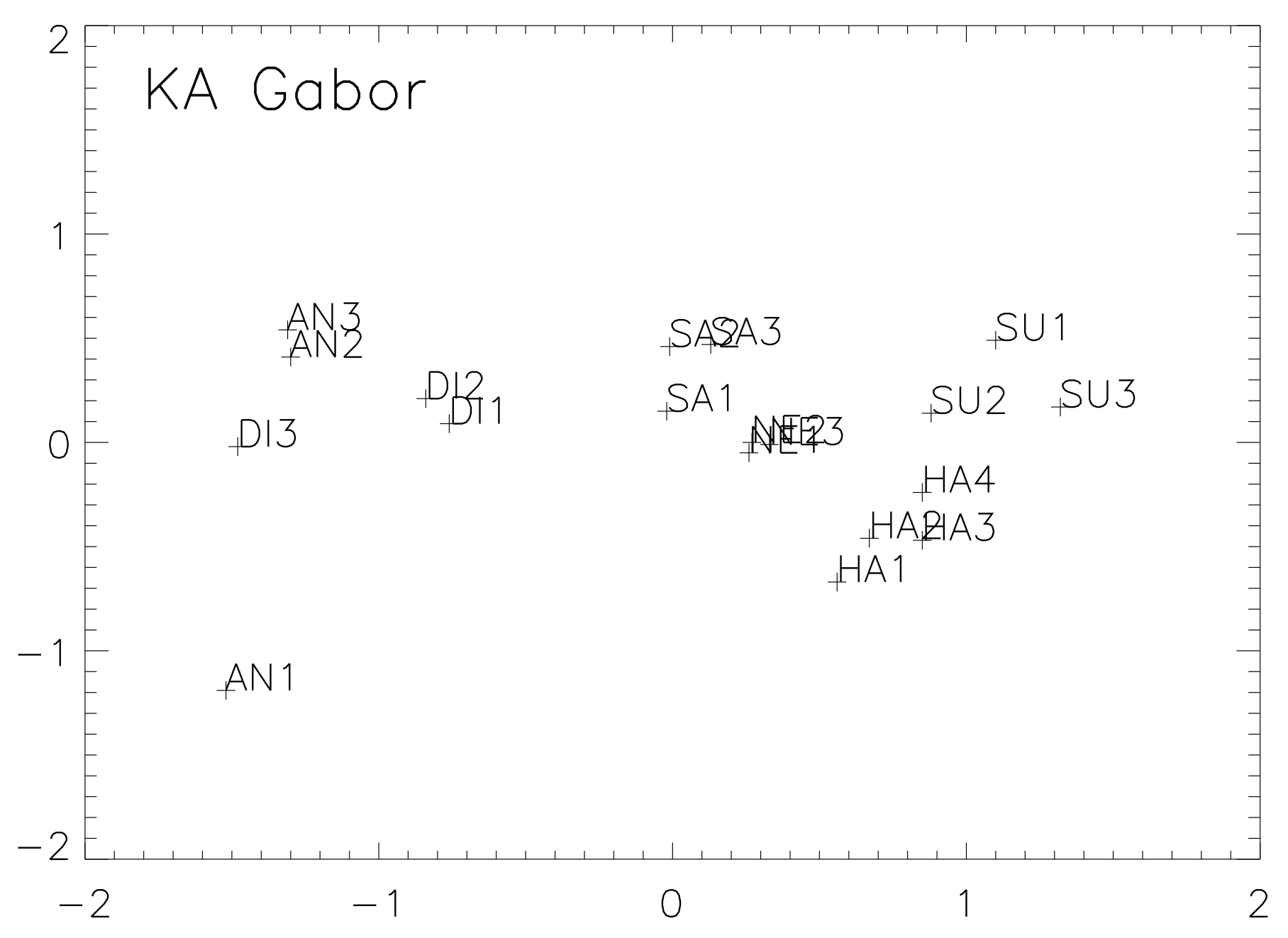}
\caption{nMDS solution spaces for Gabor and semantic rating similarities 
(Subject KA). See text for key to expression abbreviations.}
\label{Fig6}
\end{center}
\end{figure}
nMDS solutions are arbitrary up to rotation, translation and
reflection of the configuration of points. In Fig. 5 the points
have been rotated, translated, and reflected to show the agreement between
model and data.  Figs. 6 and 7 have not been treated in this way.
The most salient aspect of these plots is the relative positioning of the facial
expression clusters.
\begin{figure}[t]
\begin{center}
\includegraphics[width=.49\linewidth]{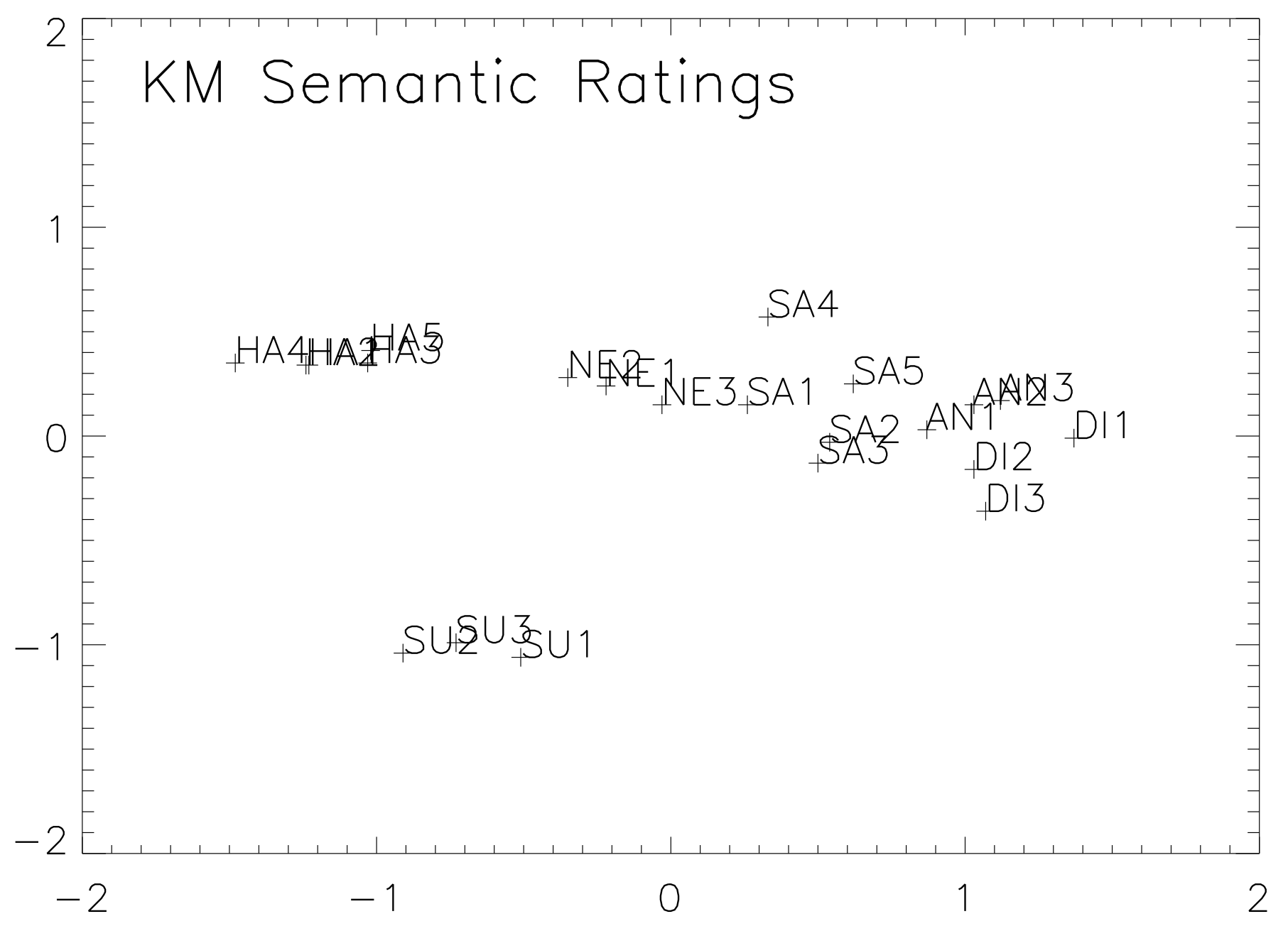}
\includegraphics[width=.49\linewidth]{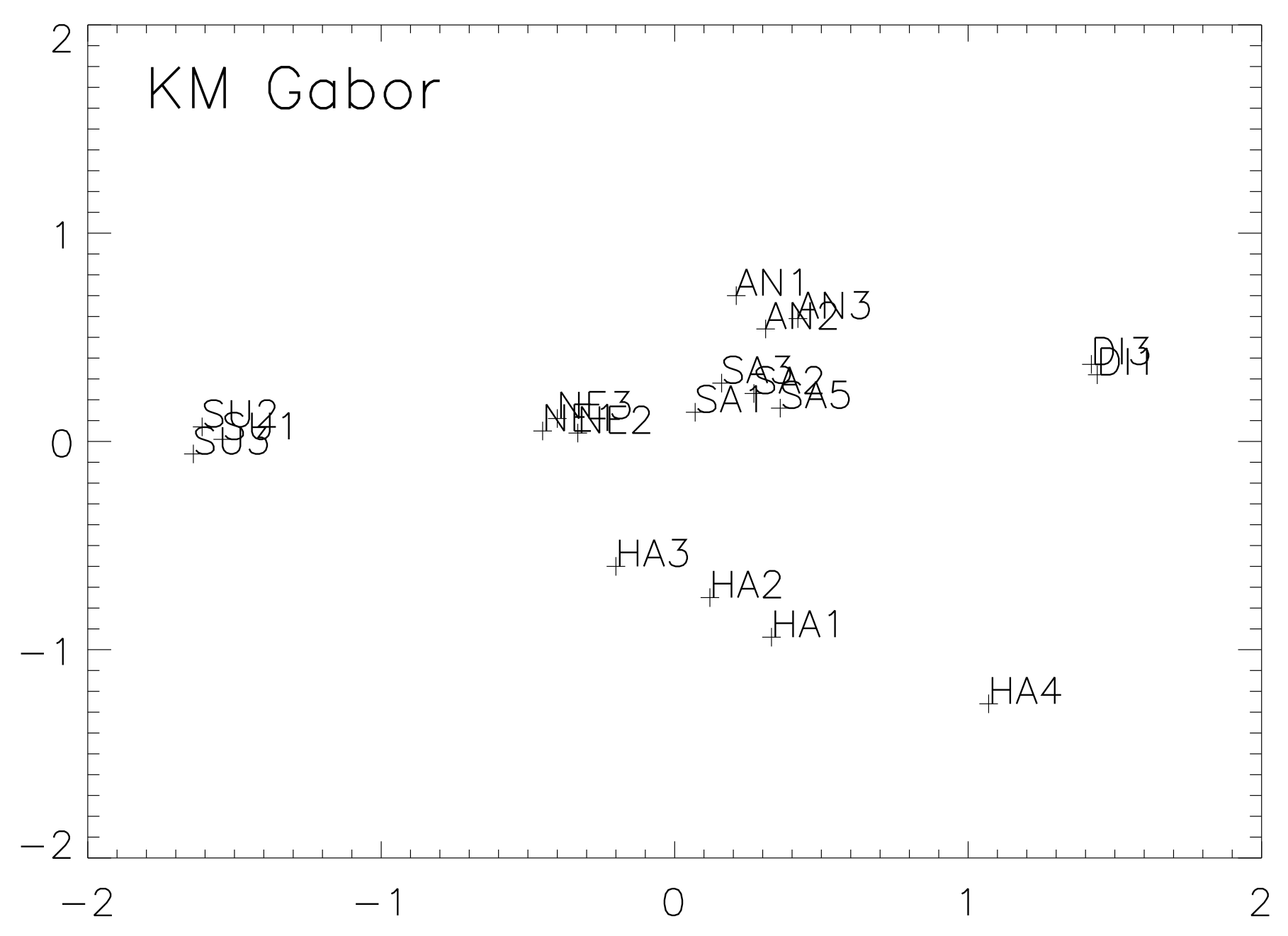}
\caption{nMDS solution spaces for Gabor and semantic rating similarities 
(Subject KM). See text for key to expression abbreviations.}
\label{Fig7}
\end{center}
\end{figure}
\section{Discussion}
Similarity values calculated using the Gabor coding and semantic
ratings showed a highly significant degree of correlation, with no
parameter fitting. Non-metric multidimensional scaling uncovered a
low-dimensional space within which Gabor-coded images cluster into
the known basic categories of facial expressions. Together, these
findings show that this representation scheme extracts adequate image
information to code facial expressions. Using this image
coding method and a multilayer perceptron classifier a facial expression
recognition system has been built \cite{Zhang}.\footnote{Not long after these results were published, we demonstrated similar performance using a more constrained analysis based on linear discriminant analysis applied to the Gabor coded images \cite{PAMI}.} Two sets of
experiments were run, one excluding fear expressions. The agreement between the model and data is higher when we exclude fear from the comparison. Fear ratings are more variable than for the other expression categories, suggesting that fear is either more difficult for our expressers to pose, or for the viewers to recognize.

Interestingly, the low-dimensional spaces for ratings data and
Gabor-coded image data are similar. One axis (nearly horizontal in Fig. 5)
corresponds to the degree of pleasantness (happy vs. anger and
disgust) in the expression.  A roughly orthogonal dimension
corresponds to the level of arousal shown by the face (surprised
vs. sad).  We observed this configuration for all of the expressers
studied (except NM, where the data is erratic). Deviations from this
general arrangement visible in Figs. 6 and 7 are typical of other nMDS
results that we do not show here.

The Gabor similarity showed a higher degree of correlation with the
data than did a geometry-based control.  Feature geometry, an
explicit and precise function of facial deformation due to expression,
does not capture any textural changes.  The addition of more grid points
could increase the performance of the geometry measure but at the cost
of increased computational complexity.  Locating the grid
points is the most expensive part of a fully automatic system
\cite{Lades}.  Moreover, the Gabor measure puts less stringent
demands on the precision of the grid position because the similarity calculation does not use the phase of the filter response. A
combined Gabor+Geometry system could have even higher performance,
but the results of \cite{Zhang} indicate the improvements are minor.

Previous studies on automatic facial expression processing classify
images into facial expression or facial action categories. The facial
images used in training or testing such systems should preferably be
pictures of pure expressions posed by trained experts.  A novel aspect
of our work is that we compare the representation with differential ratings on emotion adjectives. This procedure avoids a requirement for pure expressions. By comparing the system with human semantic rating data, we relax the relevance of expression label categories.

Why is there any agreement with psychology?  Facial expressions are
distinguished by fine changes in the shape and texture of the face. From
the standpoint of neurobiology, such changes are best represented
using the spatially localized receptive fields typical of primary
visual cortex (V1) cells. The neural systems processing facial
expressions in higher vision require access to such spatially
localized information.  Gabor wavelet functions approximately model V1
simple cell while the amplitude of the complex Gabor transform models
complex cells \cite{Daug85,JP87,PolRon}.  Hence a Gabor wavelet code
of facial expression may partially model expression coding by the
brain. Previous work by Lyons et al. \cite{ARVO} found that the Gabor
measure predicts aspects of facial similarity perception.\footnote{A more detailed account of this work was finally published in \cite{PC}.}

Finally, it is interesting that the low-dimensional structure of the
emotion adjective semantic rating data similarity space resembles that of the Gabor measure. Many studies in the psychological literature (beginning with Schlosberg \cite{Schlos}, but more recently studied by Russell \cite{Russ1,Russ2}) suggest a ``circumplex'' arrangement of the
basic facial expressions in a two-dimensional space with dimensions of
pleasantness and arousal.  We conjecture that high-level (even
semantic level) processing of facial expressions may preserve some of the topographical organizational aspects of the low-level processing by the early visual system.

\end{document}